\documentclass[letterpaper, 10 pt, conference]{ieeeconf}  % Comment this line out if you need a4paper

\IEEEoverridecommandlockouts                              % This command is only needed if 
\usepackage{times}
\usepackage{epsfig}
\usepackage{graphicx}
\usepackage{amsmath}
\usepackage{amssymb}
\usepackage{algorithm,algorithmic,multicol}

\newcommand{ \mc }[1]{\mathcal{#1}}

\newcommand{\mb}[1]{\mathbf{#1}}

\usepackage[pagebackref=true,breaklinks=true,letterpaper=true,colorlinks,bookmarks=false]{hyperref}

\title{\LARGE \bf
Belief Tree Search for Active Object Recognition
}

\author{Mohsen Malmir$^{1}$ and Garrison W. Cottrell$^{2}$% <-this % stops a space
\thanks{*This work was not supported by any organization}% <-this % stops a space
\thanks{$^{1}$Mohsen Malmir is a PhD candidate at the Computer Science and Engineering Department,
        University of California San Diego, 9500 Gilman Dr., San Diego, CA, 92093
        {\tt\small mmalmir@eng.ucsd.edu}}%
\thanks{$^{2}$Garrison W. Cottrell is a professor at the Computer Science and Engineering Department,
        University of California San Diego, 9500 Gilman Dr., San Diego, CA, 92093
        {\tt\small gary@eng.ucsd.edu}}%
}

\begin{document}

\maketitle
\thispagestyle{empty}
\pagestyle{empty}

%%%%%%%%%%%%%%%%%%%%%%%%%%%%%%%%%%%%%%%%%%%%%%%%%%%%%%%%%%%%%%%%%%%%%%%%%%%%%%%%
\begin{abstract}
Active Object Recognition (AOR) has been approached as an unsupervised learning problem, in which optimal trajectories for object inspection are not known and are to be discovered by reducing label uncertainty measures or training with reinforcement learning. Such approaches have no guarantees of the quality of their solution. In this paper, we treat AOR as a Partially Observable Markov Decision Process (POMDP) and find near-optimal policies on training data using Belief Tree Search (BTS) on the corresponding belief Markov Decision Process (MDP). AOR then reduces to the problem of knowledge transfer from near-optimal policies on training set to the test set. We train a Long Short Term Memory (LSTM) network to predict the best next action on the training set rollouts. We sho that the proposed AOR method generalizes well to novel views of familiar objects and also to novel objects. We compare this supervised scheme against guided policy search, and find that the LSTM network reaches higher recognition accuracy compared to the guided policy method. We further look into optimizing the observation function to increase the total collected reward of optimal policy. In AOR, the observation function is known only approximately. We propose a gradient-based method update to this approximate observation function to increase the total reward of any policy. We show that by optimizing the observation function and retraining the supervised LSTM network, the AOR performance on the test set improves significantly.
\end{abstract}

%%%%%%%%%%%%%%%%%%%%%%%%%%%%%%%%%%%%%%%%%%%%%%%%%%%%%%%%%%%%%%%%%%%%%%%%%%%%%%%%
\section{INTRODUCTION}
\par{Active Object Recognition (AOR) refers to the problem of predicting object label from the images while being able to change the pose of the object relative to the camera for increasing prediction certainty. A robot  rotating an in-hand object to refine its label prediction accuracy is an example of an AOR system. Ambiguity in object recognition exists because of similar views of different objects. AOR aims at finding the optimal sequence of actions which decreases the label ambiguity and improves object recognition performance in smaller number of steps. Despite its wide application and performance improvement capacity, AOR has not been applied widely and has remained secluded from main-stream computer vision progress in recent years.}

\par{Existing approaches to AOR change the sensor position to reduce the ambiguity of label prediction \cite{Aloimonos1988,Bajcsy1988,Denzler2001}. Most of these methods rely on uncertainty about object label, and use greedy best next action selection \cite{Browatzki2014} at the test time to decrease label probability entropy. A few method aim for optimal action selection at the test time using dynamic programming \cite{atanasov2014nonmyopic} or Monte Carlo planning \cite{pattenmonte}. However these methods require a model of the object, and are computationally heavy at the test time. We propose a method that that reduces optimal action selection to a simple classification of beliefs at the test time and does not require planning. We show that the proposed method generalizes well to \emph{novel views} of familiar objects, and also to \emph{novel} objects. Moreover, we show that active perception paradigm can be used to improve the accuracy of object recognition system, by selecting images that are more likely to result in higher rewards for training.}

\par{
In the first contribution of this paper, we formulate AOR as a POMDP problem and adapt a Belief Tree Search algorithm \cite{lee2007makes} to discover near-optimal values for objects poses on the training set. We infer a policy from these values, and use that to train an LSTM network to predict the best action given the current objects belief. At the test time, we use the actions suggested by this LSTM to explore the objects. We show that this supervised approach generalizes well to explore novel objects and novel views of familiar objects, and results in higher AOR accuracy compared to reinforcement learning and guided policy search methods.}

\par{In our second contribution, we derive an update rule to learn the parameters of the POMDP likelihood function with the goal of maximizing the total reward. This update rule emphasizes views of objects that will produce higher rewards in the future. We show that by retraining the likelihood function using the proposed method, the performance of the AOR system significantly improves.}

\par{
In the next section, we review some of the previous approaches to AOR. Then we present the BTS algorithm and the observation function update rule. In the results section, we report the details of the implementation of these methods and their performance on GERMS \cite{mmalmir2015} dataset. GERMS has proved to be a challenging AOR dataset, and we improve state-of-the-art performance on this dataset. The final section is the concluding remarks.  
}

\section{Literature Review}

%\par{Early approach to AOR use heuristic methods to select actions, for example to bring the object to a predefined \emph{standard} view where the object recognition accuracy is expected to be maximized. This system was developed by Wilkes and Tsotsos \cite{Wilkes1992}, in which camera position and orientation was changed to bring the object into a \lq standard\rq \, view using a robotic-arm-mounted camera. The standard view were selected subjectively to be unique among all objects in the test set. Aside from performance issues, such method clearly can not generalize to arbitrary sets of objects.}

\par{A large category of AOR models try to minimize the predicted label uncertainty through best next-action planning \cite{Schiele1998,Borotschnig2000,Callari2001,Denzler2001,Browatzki2012,Browatzki2014}. These models predict the object label probabilities using the current view, and search for the best next action that minimizes the expected entropy of object labels. In these methods, learning object appearance is performed by fitting a generative model offline, while best action selection is carried out online at the test time. Uncertainty measures such as conditional entropy and mutual information are computationally expensive to evaluate for all possible observations. Therefore these methods usually resort to approximations of these measures, which might result in poor AOR performance.}

\par{A second category of models use techniques such as REINFORCE \cite{williams1992simple} or Neurally Fitted Q-Iteration (NFQ) \cite{riedmiller2005neural} to find a good policy or action-value function for object exploration \cite{Paletta2000, mnih2014recurrent, mmalmir2015}. A parametric function that encodes object exploration policy or action-values is learned offline by using an exploration policy and collecting rewards that depend on the label prediction accuracy. The model then updates the parameters of policy or action-value function to maximize its total expected reward. These method suffer from high variance of prediction due to sampling of actions, only guarantee convergence to a local optimum and require a lot of training to explore and discover optimal sequences of actions.}

\par{More recently, deep convolutional neural networks (CNN) have been applied in AOR as a tool for modeling object appearance along with action-value prediction \cite{johns2016pairwise, Malmir2016, Haque_2016_CVPR, jayaraman2016look}. Malmir et. al trained a deep CNN using NFQ update rule \cite{Malmir2016}. In this work, a layer of Dirichlet distribution is embedded into the network for modeling the distribution of beliefs for different object-action pairs. Johns et. al used deep CNNs for entropy regression and action prediction for the set of next view points \cite{johns2016pairwise}. Finding the optimal trajectory for object inspection is then approximated by maximizing the sum of cross entropy over adjacent views pairs. Haque et al. trained LSTM networks with REINFORCE algorithm to recognize subjects from 3D point-clouds \cite{Haque_2016_CVPR}. Jayaraman and Grauman modeled object exploration policy as a neural network and trained it using classification accuracy as reward \cite{jayaraman2016look}. They found that predicting the next state of the environment based on current state and action improves the overall AOR accuracy. All these methods show improved performance over random exploration strategy and non-active methods. However they suffer from the same problem as previous methods, which is lack of guarantee of performance even on the training set.}
\par{
There are very few approaches to AOR that aim to find optimal exploration policies. Atanasov et al. \cite{atanasov2014nonmyopic} adapted an active hypothesis testing approach \cite{naghshvar2013active} for camera view-point selection for object segmentation. This approach learns a model of object appearance, and uses that for planning a sequence of actions that minimizes the cost of motor movements, object classification and view-point prediction. A dynamic programming approach is used to discover the best sequence of actions that minimizes this cost. This method depends on the representation of object appearance for efficient planning, while our method acts in the belief space and is completely independent of object representation and classification. Patten et al. \cite{pattenmonte} use Monte Carlo planning for active exploration and perception of outdoor objects. This method uses rollouts that depends on predicting the point-cloud of objects for different actions. Compared to this method, our method is more intuitive, and doesn't require a model of objects. 
}
\par{
Of special interest to this paper are works on belief tree search and Monte Carlo POMDP planning. Lee et. al \cite{lee2007makes} proposed clustering of beliefs in a belief tree search algorithm to reduce the width of the tree. DESPOT \cite{somani2013despot} uses sampling of observations to reduce the width of the belief tree for optimal action selection. POMCP \cite{silver2010monte} adapts Monte Carlo sampling and the Upper Confidence Trees algorithm \cite{kocsis2006bandit} for efficient POMDP planning. We based our method to \cite{lee2007makes} because of desirable properties such as acting on the belief space and performance guarantees on the \emph{reachable} space of beliefs.
}
\par{
An important features of the proposed approach is its freedom of maintaining an object model. Our method acts in the belief space of objects, and only requires a black box simulator for training, that returns the belief resulting from performing different actions on objects. Another important property of this method is that at the test time, the next action is predicted by a simple classification of the current belief. We show that this approach is more effective in learning object exploration policies, compared to reinforcement learning and actor-critic methods. Next section describes the proposed approach in more details.}

\section{Monte Carlo Belief Trees}
\par{
Active object recognition can be formulated as a POMDP problem denoted with tuple $<\mc{S},\mc{A},\mc{O},\mc{T},\mc{P},\mc{R},\gamma>$, where $\mc{S}$ is the set of states (object labels), $\mc{A}$ is the set of actions for object examination, and $\mc{O}$ is the set of observations (captured images of objects). We don't perceive the identity of object directly but rather collect information about it through observations. The transition function $\mc{T}$ marks the transition between different states and the observation function $\mc{P}(s,a,o)$ relates the observations to different object identities through $\mc{P}(s,a,o) = Pr(o | s,a)$ which is the probability of observing $o$ after taking action $a$ when the object labels is $s$. The reward function $\mc{R}(s,a)$ determines the reward for taking action $a$ when the object label is $s$. Finally, $\gamma$ is the reward discount factor.
}
\par{
Note that in this POMDP, the transition function $\mc{T}$ reduces to the identity function. The observation space on the other hand includes all images of objects, and is prohibitively large to apply value iteration techniques \cite{pineau2003point}. Another approach is to use Monte Carlo planning , which builds a search tree for one-step action selection \cite{silver2010monte}. However this method suffers from the curse of history, which is exacerbated by high dimension observations. We seek to find a solution to the AOR POMDP that compactly represents history, and allows tractable search for finding the optimal actions.
}

\par{
It has been shown that solution to POMDP can be found by solving the equivalent \emph{belief} MDP, where the belief $b \in \mathbb{R}^{|\mc{S}|}$ denotes the posterior probability of states given the observation history,
\begin{align}
b^t(s) &= Pr(s | a_{0:t-1},o_{0:t-1}) \\
\sum_{s \in \mc{S}} b(s) &=1,\;\; b(s) \geq 0 \; \; \forall s \in \mc{S}.\nonumber
\end{align}
In belief MDP, states represent the posterior probabilities of POMDP states given the action-observation history. For this MDP, the transition between beliefs given action $a$ is defined as,
\begin{align}
\mc{T}&^{MDP}(b,a,b') \nonumber \\ 
&= Pr(b' | b,a) \nonumber \\
\label{eq:blftransition}&= \sum_{o \in \Omega(b,a,b')} \sum_{s,s'\in \mc{S}} b(s) \mc{T}^{POMDP}(s,a,s') \mc{P}(s',a, o)
\end{align}
where $\Omega(b,a,b')$ denotes the set of observations that can result in changing beliefs from $b$ to $b'$,
\begin{align}
\Omega&(b,a,b') = \{ o\in \mathcal{O} | \nonumber \\
&b'(s')= \frac{ \mc{P}(s',a,o) \sum_{s\in\mathcal{S}} \mc{T}^{POMDP}(s,a,s') b(s)}{Pr(o | a,b)}\}
\end{align}
The belief MDP reward is defined by calculating the expected reward over states,
\begin{align}
R(b,a) = \sum_{s \in \mathcal{S}} b(s) \mathcal{R}_s^a.
\end{align}
And finally, given an initial belief $b$, action $a$ and observation $o$ the updated belief $b'$ is calculated by,
\begin{align}
\label{eq:beliefupdate}
b'(s') &= Pr(s' | b,a,o) \nonumber \\ 
&= \frac{\mc{P}(a,s',o) \sum_{s\in\mc{S}} b(s)\mc{T}^{POMDP}(s,a,s')}{Pr(o|a,b)}.
\end{align}
}

\par{
Now that we defined the equivalent belief MDP, we aim at solving the planning problem using \emph{Belief Tree Search} algorithm. This algorithm constructs a search tree for a given belief $b$, where different branches represent actions and the resulting observations. Each node in the tree represents a belief about the underlying POMDP states and edge captures an action and the resulting observation. The algorithms starts with $b$ as the root and exhaustively performs all action to collect new observations and form new beliefs. These beliefs are then added as children of the root and the process iterates with new beliefs until stop states are reached in the leaves. The values are then backtracked from the leaves to the root to estimate the value of $b$. The belief tree search is used in online planning for POMDPs where the dynamics of the environment are known. However we adapt it to calculate the optimal values of images for the training set.
}

\par{The plain belief tree search algorithm is computationally intractable to use for AOR belief MDP, since it examines all observations after each action. Instead we adapt the algorithm in theorem 1 of \cite{lee2007makes}, which sacrifices optimality of the predicted values in exchange of computational tractability. This algorithm utilizes the smoothness of the optimal value function to cluster the belief space. More specifically, for a given $\delta > 0$, each node in the tree represents the approximate value of beliefs that are in $\delta$-neighborhood of $b'$, which we represent as $\delta(b)$. By clustering the belief space, the width of the belief tree decreases to a manageable size. Another approximation in this algorithm is that this algorithm calculates the approximate optimal value of a given belief $b$ only in $\mc{R}(b)$, which is the \emph{reachable space} of $b$. The reachable space is defined as the set of beliefs that are reachable by arbitrary sequences of actions from $b$. Finally, the algorithm utilizes the discount factor $\gamma$ to limit the height of the belief tree.}

\par{We adapt this algorithm to find an approximately optimal value for each image in the training set. The algorithm is depicted in pseudo code style in algorithm \ref{alg:MOCBET}. Using these values, training an active object recognition system reduces to a supervised learning and knowledge transfer problem. In each node of the tree, the algorithm expands all actions and receives the new observations. New beliefs are then calculated using (\ref{eq:beliefupdate}). For each new belief $b'$, if there is an already expanded belief $b_0$ in that height of the tree for which $b' \in \delta(b_0)$, the algorithm sets the value of $b'$ equal to $b_0$ and backtracks. Otherwise, the search continues in the children of $b'$. 
}

\par
\begin{figure}
\centering
\includegraphics[width=0.35\textwidth]{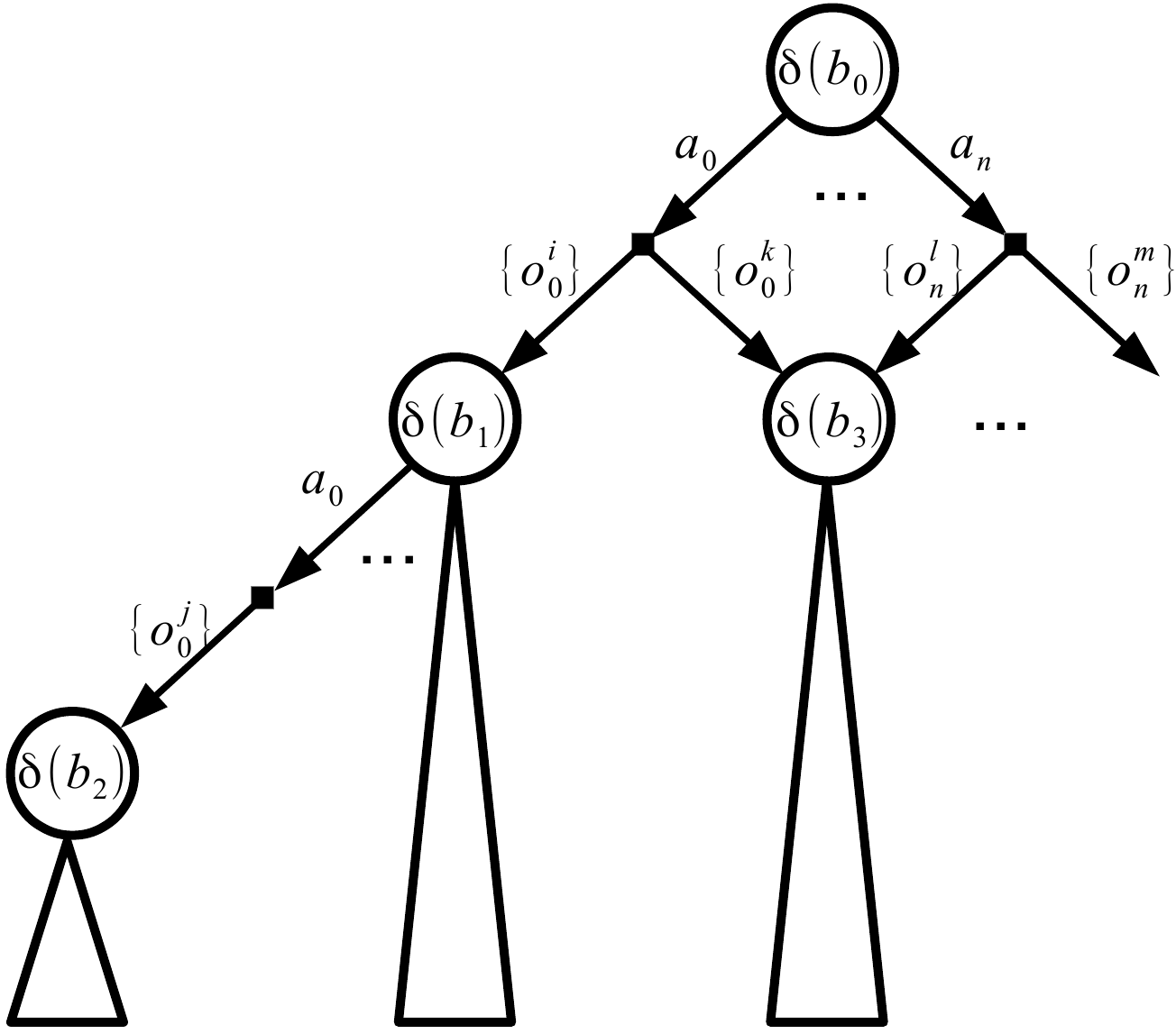}
\caption{Belief Tree Search algorithm. Each node in the tree represents the value of $\delta$-neighborhood} of a belief with some error. A set of observations may lead to visiting the same belief after taking an action.
\label{fig:belieftree}
\end{figure}

{
Our algorithm builds a belief tree over $\mb{R}(b_0)$ by sampling from images in the training set and maintaining a $\delta$-packing of $\mb{R}(b_0)$ at each level of the tree. A belief tree with root $b_0$ denotes all the possible actions and observations that are encountered while inspecting an object with the initial belief $b_0$. A belief tree captures all the possible actions and observations, construction of full belief tree is prohibitive in case of active object recognition because the size of the observation space is extremely large. One modification that we made to algorithm \ref{alg:MOCBET} compared to theorem 1 of \cite{lee2007makes} is that at the root node, we calculate the value of $\delta(b_0)$. This is to reduce the overfitting of values to specific beliefs. In our algorithm, if two beliefs are very similar but result in vastly different rewards, that should be considered in calculating the value of $b_0$. In AOR this happens when the classifier is uncertain about some examples then their beliefs are close to each other and this should be reflected in their calculated values. Figure \ref{fig:belieftree} depicts the proposed algorithm in graphical form.
}

\begin{algorithm}[h]
  \caption{Belief Tree Search}
  \label{alg:MOCBET}
  
    \begin{algorithmic}
      \scriptsize
          \STATE \textbf{Belief Tree Search}\text{(Belief $b_0$, radius $\delta$, max height $h$)}
          \FOR{All Images $o_i$ in training set}
                \STATE $b_i\leftarrow $ Classify($o_i$)
                \IF {$b_i \in \delta(b_0)$}
                	\STATE Expand($o_i,b_i,0$)
                \ENDIF
      	  \ENDFOR
    \end{algorithmic}

    \begin{algorithmic}
      \scriptsize
      	   \STATE \textbf{Expand}\text{(Image $o$, Belief $b$, level $i$)}
	   			\IF {$i = h$}
					\RETURN
				\ENDIF
           		\FOR{action $a \in \mc{A}$}
           	 		\STATE Image $o_a \leftarrow$ Simulate-Action($o,a$)
             		\STATE Belief $b_a\leftarrow $ Transition($b,a,o_a$)
             		\IF{ $b_a \in \delta$-neighborhood of any $b'$ belief at level $i+1$}
                 		\STATE $V(b_a) \leftarrow V(b')$
					\ELSE
             			\STATE Add $b_a$ to nodes in level $i+1$
             			\STATE Expand($o_a, b_a, i+1$)
             		\ENDIF
					\STATE add ($o_a,b_a$) to children of $b$
           		\ENDFOR
           		\RETURN
           \end{algorithmic}
\end{algorithm}

\par{
Theorem 1. provides some guarantee on the optimality of the values found for images in algorithm (\ref{alg:MOCBET}).\\
\textbf{Theorem1.} For a given maximum error $\epsilon$, and the optimal value function $V^*(b)$, algorithm (\ref{alg:MOCBET}) finds the value of a given belief $b_0$ such that $|V^*(b_0) - V(b_0)| \leq \epsilon$ by setting the parameter values as, 
\begin{align}
\delta &= \frac{\epsilon (1-\gamma)^2 }{2 R_{max}} \\
h &= \log_{\gamma} \frac{(1-\gamma)\epsilon}{2R_{max}}
\end{align}
\textbf{Proof}. See supplementary materials.
}

\subsection{Optimizing Observation Function}
\par{
It is very important to understand that the observation function in AOR POMDP is an approximation to the actual likelihood values of object views under different classes, which depends on the model that one fits to the data to model the likelihood. For example Borotsching et. al use Gaussian mixture on the eigenspace to model the likelihood of images from the view sphere of object under different classes \cite{Borotschnig2000}, while Malmir et. al approximate the observation function using a deep convolutional networks \cite{Malmir2016}. Calculating the observation function value for an image usually requires feature extraction from the image and density estimation for different classes. We assume a parametric observation function, which is a function of $\phi$,
\begin{align}
\mc{P}(s,a,o ; \phi) = Pr(o | s,a; \phi)
\end{align}
Different values of parameters $\phi$ changes the approximation to the observation function. One may \emph{improve} the observation function by using a different estimation of $\phi$, which changes the feature extraction or density estimation. The improved observation function then results in a POMDP with different environment dynamics. In the ideal case the observation function value given an image is 1 for the correct object label and 0 for other labels, in which case the POMDP reduces to a trivial MDP.}
\par{
For a given observation function, theorem 1 finds the image values for policies arbitrarily close to the optimal policy. However these values depend on the POMDP dynamics, e.g. the observation function. We propose to improve the observation function by increasing the total reward collected by the near-optimal policy. For any policy $\pi$ and observation function $\mc{P}$, the total reward $\rho$ is defined as,
\begin{align}
\rho(\pi,\mc{P}) &= E \left  \{ \sum_{t=1}^\infty \gamma^t r_t | s_o,\pi  \right \} \nonumber \\
&=\int_{b \in \mathcal{B}} d^{\mathcal{P},\pi}(b) \sum_{a  \in \mathcal{A}} \pi(b,a) R(b,a) db
\end{align}
Where $\mathcal{B}$ is the $|\mc{S}|$-dimensional belief simplex. Changing the observation function parameters may result in increased likelihood of images for the corresponding object label. Theorem 2 presents a gradient ascent update rule to the parameters of observation function, with the goal of increasing the total reward.
}
\par
{
\textbf{Theorem 2.} Given a policy $\pi$ and the corresponding value function $V^{\pi}$ the gradient of the total reward $\rho(\pi,\mc{P})$ with respect to the parameters of the observation function $\mathcal{P}(;\phi)$ is given by,
\begin{align}
\label{eq:theorem2}
&\frac{\partial}{\partial \phi} \rho(\mathcal{P},\pi) \nonumber \\
&=\int_b d^{\mathcal{P},\pi}(b) \sum_a \pi(b,a) \int_{b'}  \sum_{o \in \Omega(b,a,b')}
\sum_{s,s' \in \mathcal{S}} \nonumber \\
& \; \; \; \; b(s) T(s,a,s') V^{\pi}(b') \frac{\partial}{\partial \phi} \mathcal{P}(o|s',a) \text{d}b' \text{d}b
\end{align}

\textbf{Proof.} See the supplementary notes.
}
\par{
Intuitively speaking, the update rule in (\ref{eq:theorem2}) weights each parameter by the value of the belief that is reached by observing the corresponding $o$. Changing the observation function parameters $\phi$ changes the belief MDP dynamics as the transition probabilities in (\ref{eq:blftransition}) depend on $\mc{P}$. After updating the observation function, a new belief MDP is reached which can be solved approximately using theorem 1. In practice evaluation of (\ref{eq:theorem2}) is computationally intractable. In the results section, we describe a simple procedure for updating the observation function based value weighted updates. 
}

\par{

\section{Results}
\subsection{BTS for Training Set}
\par{
In this section, we adapt the proposed method in algorithm (\ref{alg:MOCBET}) for active recognition of GERMS \cite{mmalmir2015}. This is a medium size dataset with $\sim120K$ images of 136 different object collected by a robot. The robot grabs each object with 10 different orientations and examines the object by rotating them in front of the camera. The goal is to recognize object for 4 test orientations, given the other 6 in-hand orientations. GERMS is proved to be a challenging dataset since separation of objects in this dataset requires extraction of small visual cues and fine categorization. %See Figure \ref{fig:cancer} for examples of objects that are different in color of the eyes.
}
\par{
We extract visual features from GERMS images using ResNet deep CNN model \cite{He_2016_CVPR}. A softmax layer is trained on top of these features to predict the object label. Then we convert train and test images into belief vectors, and train our AOR method in the belief space. We normalize the output of the softmax layer for each class to sum to 1 over all GERMS images and use that as the observation probability. This is to ensure that the deep CNN outputs the likelihood and to maintain the integrity of (\ref{eq:blftransition}) and (\ref{eq:beliefupdate}).
}
\par{
We calculate the likelihood of each image and use algorithm \ref{alg:MOCBET} to calculate the value of the near-optimal policy for each image in the train set. In order to use these values in planning, \cite{lee2007makes} proposes a sampling approach that repeatedly executes algorithm \ref{alg:MOCBET} for different simulations and augments the tree with newly discovered beliefs and finally uses the action-values of the root of the resulting tree for planning. The proposed algorithm in \ref{alg:MOCBET} is similar to the proposed approach in \cite{lee2007makes} however we make use of the similar belief vectors in the dataset to run the simulations. We found the action-values of the root of the tree to be very effective for AOR. 
}
\par{
After we extract the action-values for each training image, we transfer the knowledge of these action-values to the test set. We compare three approaches for learning policy for these action-values. In the first and second approaches, we use \emph{Neurally Fittred Q-learning} (NFQ) \cite{riedmiller2005neural} and \emph{Actor Critic} (AC) \cite{peters2008reinforcement} approaches, guided by a probabilistic policy that uses the action-values from BTS. We show that guiding these two approaches results in slight improvement of average performance on the test set, compared to the plain version. In the third approach, we use an LSTM network to learn to predict the best action from the BST action-values. We show that the LSTM network is superior in performance to NFG and AC approaches.
}

\subsection{Guided Neurally Fitted Q-learning}
\par{
Neurally Fitted Q-learning (NFQ) trains a neural network to predict the action values using the reward signal from the environment \cite{riedmiller2005neural}. This algorithm has been successfully applied reinforcement learning benchmarks \cite{riedmiller2005neural}, playing Atari games \cite{mnih2015human} and active object recognition \cite{mmalmir2015}. At the heart of this approach is an iterative update rule for the network parameters $\theta$,
\begin{align}
\label{eq:NFQ}
\theta_{t+1} \leftarrow \theta_t + \alpha \frac{\partial}{\partial \theta}(R_t + \gamma \max_a Q(s_{t+1},a) - Q(s_t,a_t))^2
\end{align}
where the network outputs action-values $Q(s,a)$ for action $a$ in state $s$. In the above, the gradient operator on the right side applies only to $Q(s_t,a_t)$. We previously observed that the plain NFQ algorithm may fail to discover optimal policies for active object recognition \cite{Malmir2016}. Instead, we employ the $n$-step extension of this algorithm, proposed in \cite{mnih2016asynchronous}, in which the update rule in (\ref{eq:NFQ}) is applied to action sequences of length $n$. The $n$-step NFQ speeds up learning by updating $n$ action-values in each iteration, compared to a single action-value update in the original NFQ. All experiments reported here are obtained using $10$-step sequences of actions.
}
\par{
We further improve the performance of NFQ by employing the \emph{importance sampling} framework for policy improvement \cite{jie2010connection}. The idea behind this approach is to use an auxiliary policy $\pi_{\theta'}$ to acquire sequences of actions and states $\tau = \{s_0,a_0,s_1,a_1,\ldots\}$, and update the parameters of target policy $\pi_\theta$ using these sequences. In order to obtain an unbiased estimator, the gradients of the policy are weighted by their importance defined as,
\begin{align}
\label{eq:GNFQ}
\frac{P(\tau | \pi_\theta)}{P(\tau | \pi_{\theta'})} = \frac{\prod_j \pi_\theta(a_j|s_j)}{\prod_j \pi_{\theta'}(a_j|s_j)}
\end{align}
Where $P(\tau|\pi)$ is the probability of sequence $\tau$ under policy $\pi$, and $\pi(a|s)$ is the probability of action $a$ in state $s$ under policy $\pi$. We implement the guided NFQ (GNFQ) by drawing sequences of actions from an stochastic policy acquired by performing softmax on BTS action-values. The gradients in (\ref{eq:NFQ}) are then multiplied by their importance in (\ref{eq:GNFQ}) and applied to the network. Figure \ref{fig:NFQ} shows the comparison of mean AOR performance of NFQ and GNFQ, on the GERMS test data. For both approaches, we show $1-\sigma$ interval of performance in shaded area. For comparison, we report the performance of random policy (RND), where at each time step, a new action is taken to explore the object. We see that the plain NFQ algorithm fails to perform any better than RND, while GNFQ performs better than random. The advantage of GNFQ over NFQ and RND is most significant for the first action, and it gradually decreases over the next four actions. After the last action where the majority of evidence has been accumulated, all three methods perform similar.
}
\begin{figure}[ht!]
\centering
\includegraphics[width=0.35\textwidth]{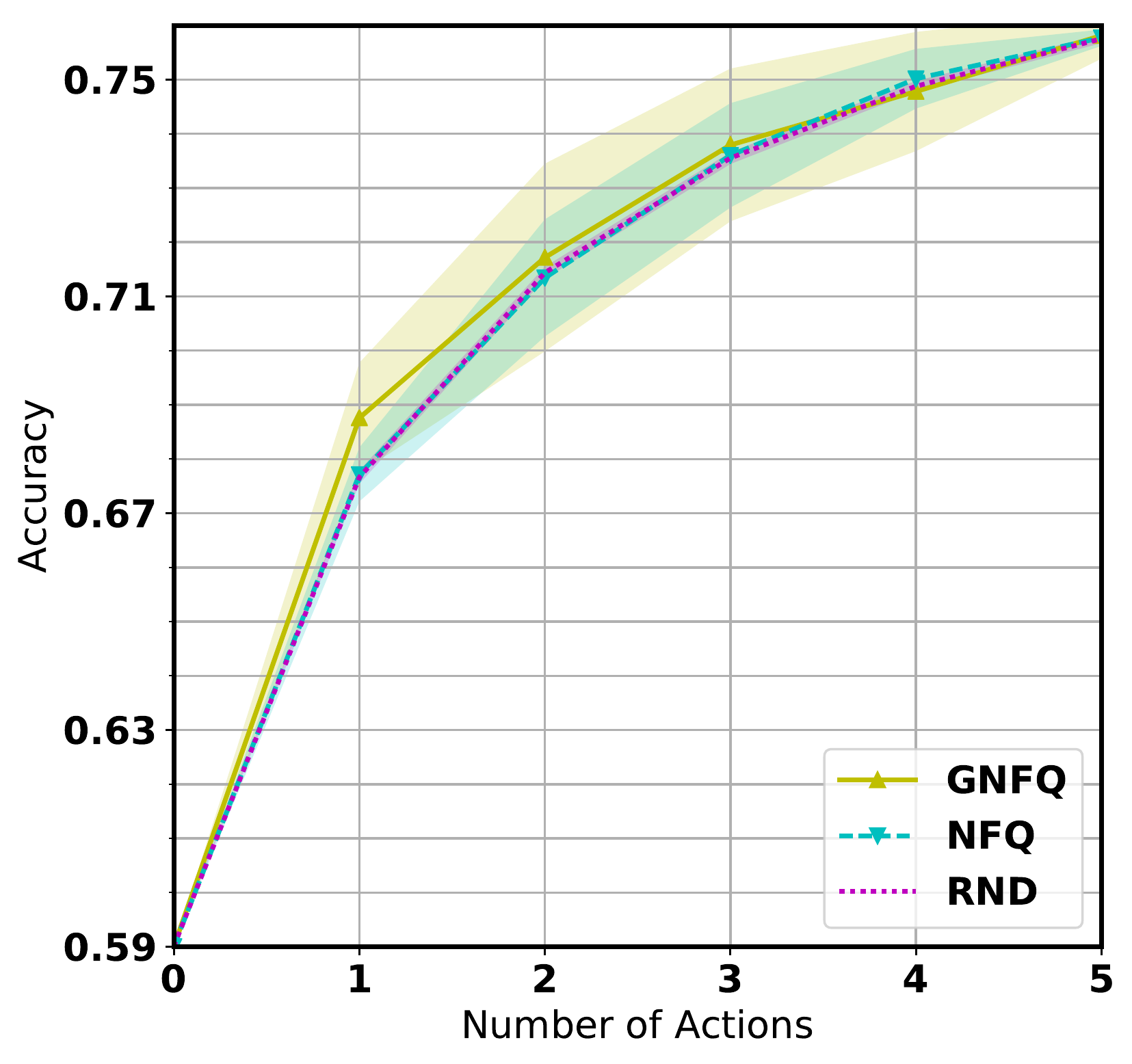}
\caption{Comparison of AOR performance for NFQ, guided NFQ and random policies. The shaded area shows 1-$\sigma$ of the mean performance.}
\label{fig:NFQ}
\end{figure}

\subsection{Guided Actor-critic}
\par{
Actor-critic is a policy learning method which updates the policy parameters using gradients of the expected reward \cite{peters2008reinforcement}. To reduce the variance of gradient estimation, predicted value is decreased from the reward,
\begin{align}
\label{eq:AC1}
\theta_t \leftarrow \theta_t + \frac{\partial}{\partial \theta} \log \pi_\theta(a_t|s_t) (R-V_\psi(s_t))
\end{align}

\begin{align}
\label{eq:RV}
\psi_t \leftarrow \psi_t + \frac{\partial}{\partial \psi} (V_\psi(s_t)-R)^2
\end{align}
where $R$ is the collected reward at time $t$, $V_\psi(s_t)$ is the predicted value for $s_t$ parameterized by $\psi$, and $\pi_\theta(a_t|s_t)$ is the probability that policy $\pi_\theta$, parameterized by $\theta$, assigns to action $a_t$ in state $s_t$. In figure \ref{fig:AC}, we compare the AOR performance of $10$-step actor-critic method with (ACG) and without (AC) guiding. We used the same guiding scheme as described above, by multiplying the gradient terms of (\ref{eq:AC1}) and (\ref{eq:RV}) by their importance (\ref{eq:GNFQ}). We see that plain AC fails to perform better than random, while ACG shows higher performance than RND after the second and third actions. 
}
\begin{figure}[ht!]
\centering
\includegraphics[width=0.35\textwidth]{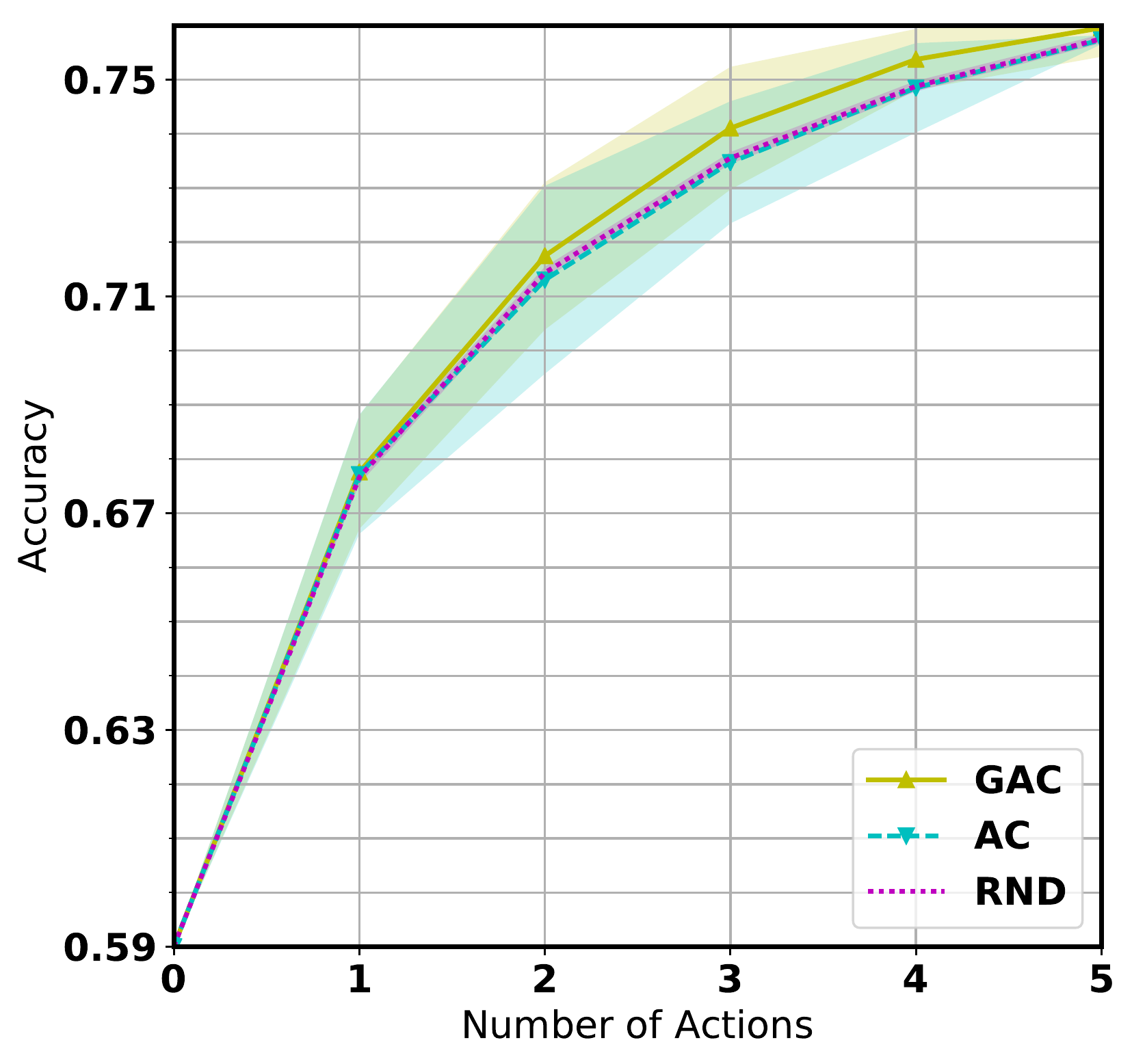}
\caption{Comparison of AOR performance for actor-critic, guided actor-critic and random policies. The shaded area shows 1-$\sigma$ of the mean performance.}
\label{fig:AC}
\end{figure}

\subsection{Supervised Learning of Action Values}
\par{
To transfer the knowledge of near-optimal values from training to test set, we train a neural network to directly predict the action-values given a belief. We use a stack of 3 LSTM layers with 128 units in each layer followed by a softmax layer that predicts action values (see Figure \ref{fig:model}). The training sequences are produced by following a probabilistic policy derived from BTS action-values. For each belief vector, the target action is the one with the highest action-value. We found that it is crucial to the performance of AOR to use sequences of data for supervised action prediction. Figure \ref{fig:LSTM} compares the performance LSTM and random policies. LSTM has a clear advantage in performance to the random policies. Moreover, the variance of the learned policies is significantly smaller compared to actor-critic and NFQ methods. Figure \ref{fig:all} compares the average performance of LSTM with previous methods. We see that supervised learning for action-value prediction is clearly superior to the policy learning methods. Table \ref{table:perf} shows the comparison of all methods in more details.
}
\begin{figure}[ht!]
\centering
\includegraphics[width=0.49\textwidth]{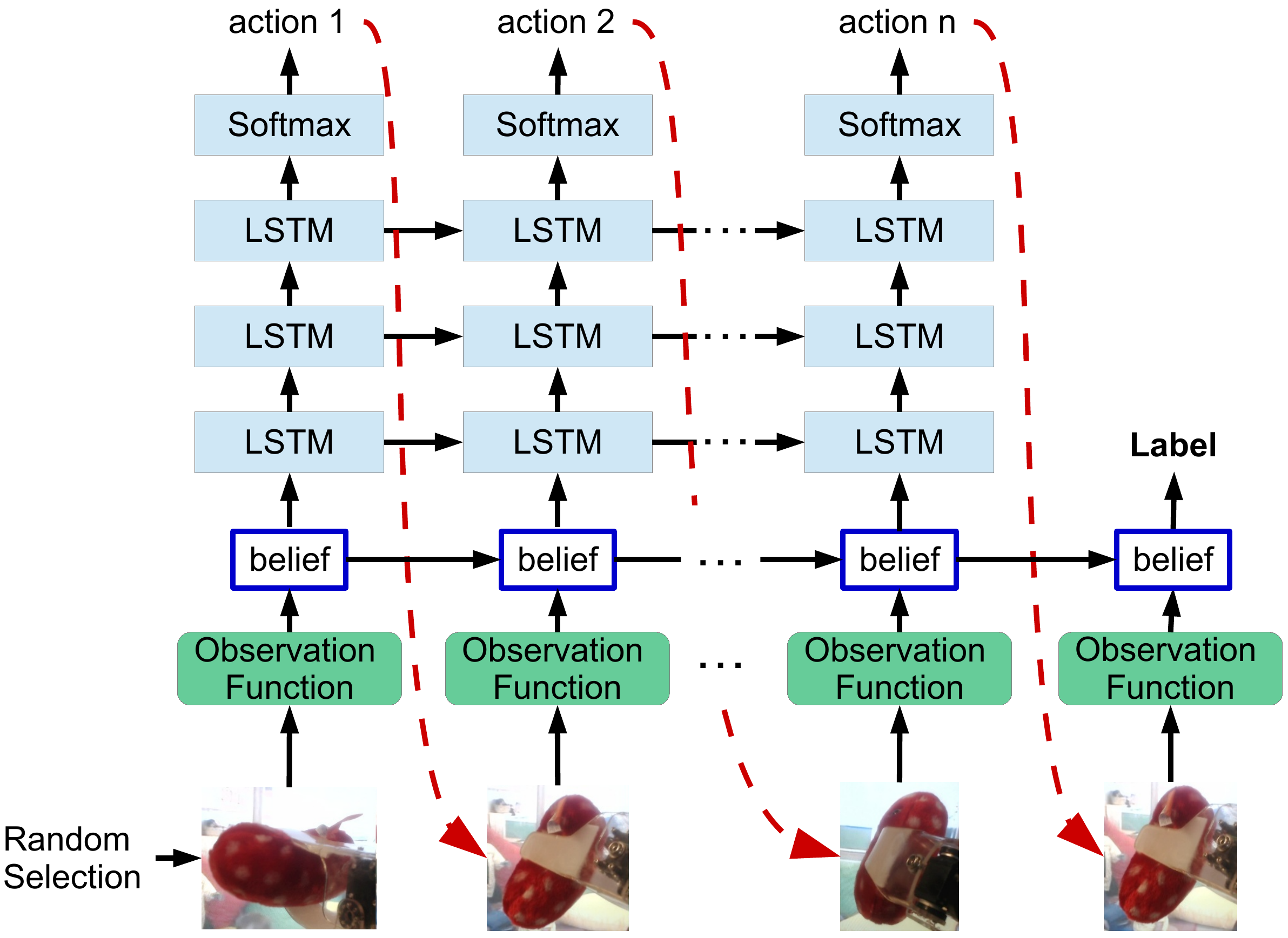}
\caption {Supervised learning of actions using LSTM network.}
\label{fig:model}
\end{figure}

\begin{figure}[ht!]
\centering
\includegraphics[width=0.35\textwidth]{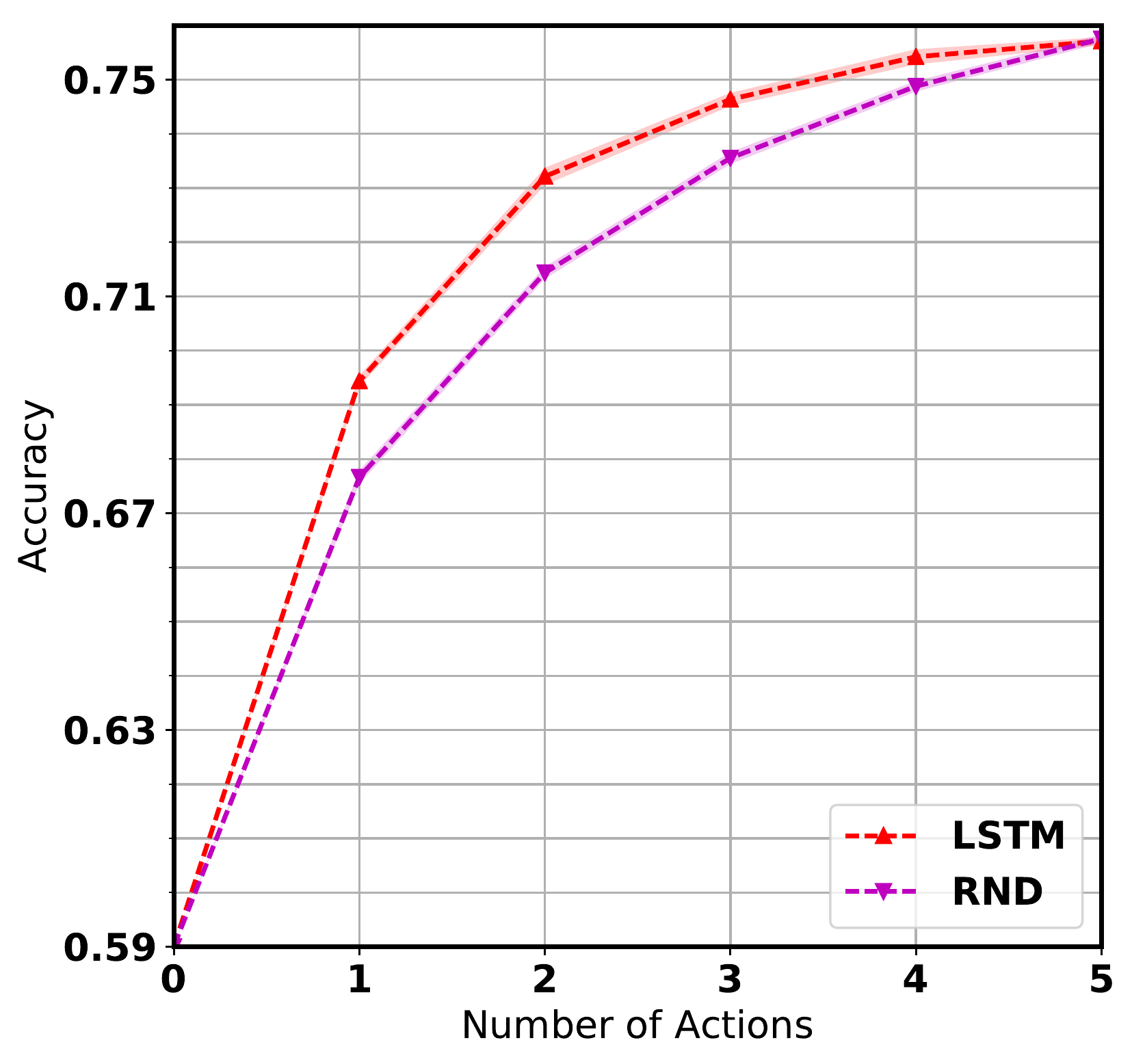}
\caption {Comparison of AOR performance for LSTM and random methods. The shaded area shows 1-$\sigma$ of the mean performance.}
\label{fig:LSTM}
\end{figure}

\begin{figure}[ht!]
\centering
\includegraphics[width=0.35\textwidth]{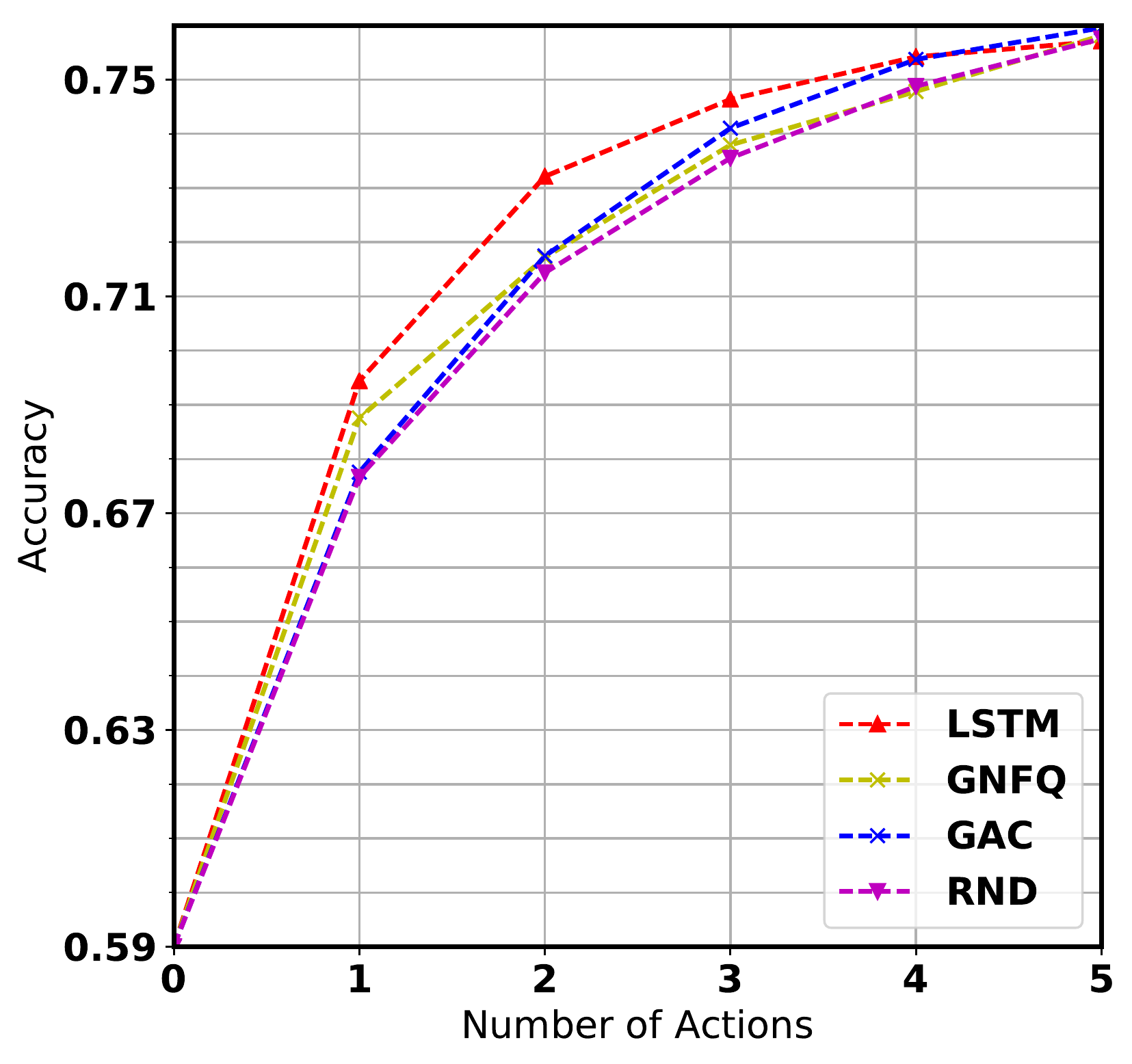}
\caption{Comparison of AOR performance for LSTM, AC, NFQ and RND policies.}
\label{fig:all}
\end{figure}

\begin{table}[h]
\caption{AOR performance comparison on the GERMS test data based on the number of actions.}
\label{table:perf}
\begin{center}
\begin{tabular}{|c||c|c|c|c|c|c|}
\hline
method & 0 & 1 & 2 & 3 & 4 & 5 \\
\hline
RND&0.590&0.677&0.714&0.736&0.749&0.758\\
\hline
AC&0.590&0.677&0.713&0.735&0.748&0.757\\
\hline
ACG&0.590&0.678&0.717&0.741&0.754&0.760\\
\hline
NFQ&0.590&0.677&0.713&0.736&0.750&0.758\\
\hline
NFQG&0.590&0.688&0.717&0.738&0.748&0.758\\
\hline
LSTM&0.590&0.694&0.732&0.746&0.754&0.757\\
\hline
LSTM-i2&0.614&0.715&0.751&0.769&0.778&0.785\\
\hline
LSTM-i3&\textbf{0.617}&\textbf{0.718}&\textbf{0.758}&\textbf{0.776}&\textbf{0.790}&\textbf{0.793}\\
\hline
\end{tabular}
\end{center}
\end{table}

\subsection{Generalization to Novel Objects}
\par{
In this section, we test the generalization of the proposed AOR method to novel object. The goal of this experiment is to understand how much object inspection knowledge is transferrable, amongst different objects of GERMS. For this purpose, we use 60\% of objects in GERMS for training the LSTM method, and used the rest of objects for testing. The results averaged over 20 different experiments are shown in figure \ref{fig:novelobj}. Overall, the variance of results is high because of the large variations in the accuracy of GERMS objects. We see that the proposed method achieves better performance compared to random. The random strategy has difficulty in finding the informative moves, compared to the novel views experiment.
}

\begin{figure}[ht!]
\centering
\includegraphics[width=0.35\textwidth]{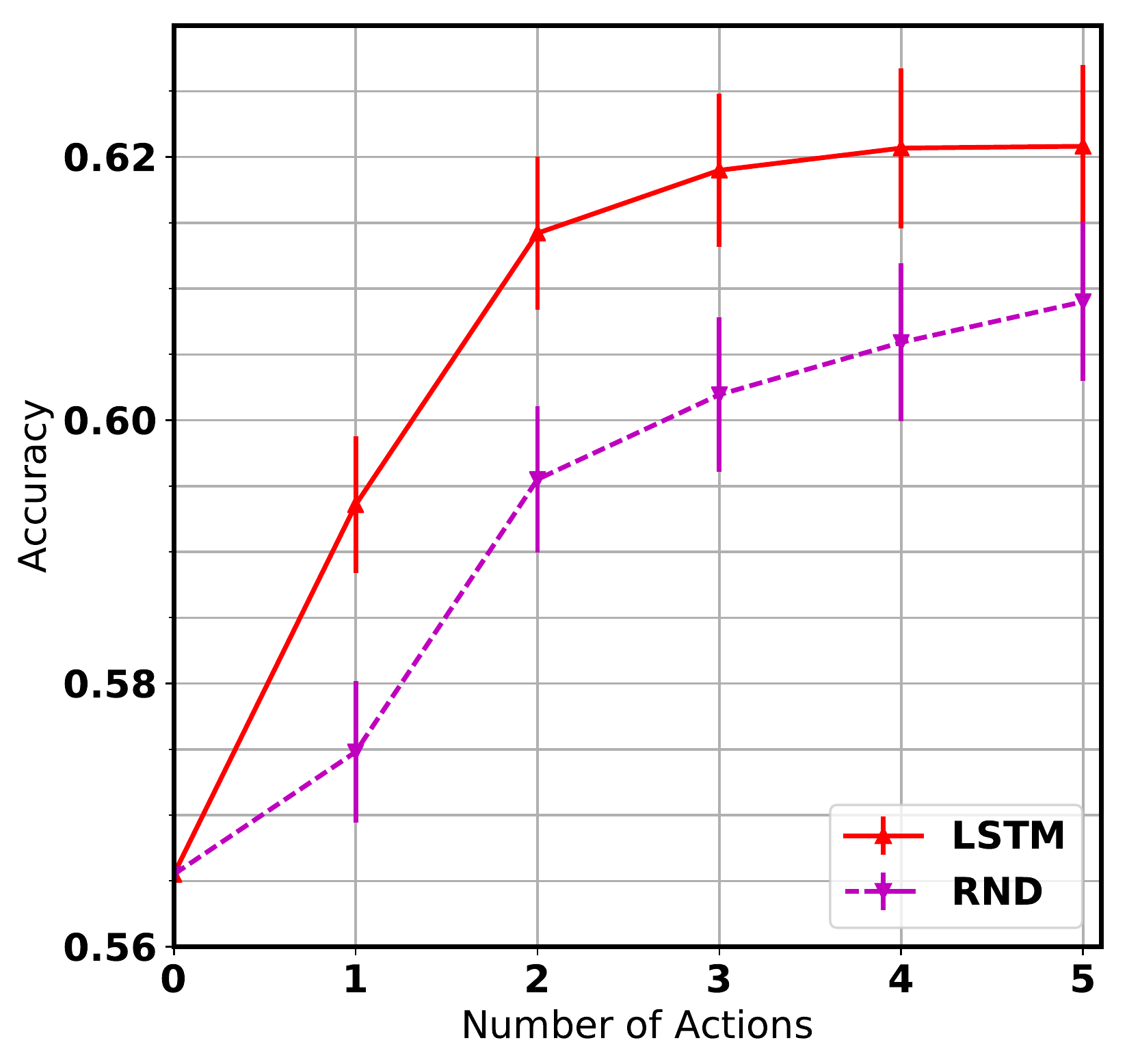}
\caption{Generalization of LSTM model to novel objects. The reported numbers are mean$\pm$ standard error.}
\label{fig:novelobj}
\end{figure}

\subsection{Improving Observation Function}
\par{
In this section, we retrain the observation function using the proposed gradient update rule in (\ref{eq:theorem2}). In order to implement this update rule, we adapt a sampling strategy by generating a set of rollouts using policy $\pi$ derived from BTS action-values. Let $b_i$ denote the belief vector corresponding to image $I_i$. Let $\{b_j,a_j\}, j=1,2,\ldots,n$ denote the set of beliefs and actions that resulted in $b_i$ in the rollouts. The retraining weight of $I_i$ is then calculated using a sample average of (\ref{eq:theorem2}) on these rollouts,
\begin{align}
\label{eq:sampleW}
Weight(b_i) \propto \sum_{j=1}^n \pi(b_j,a_j) b_j(s) V^\pi(b_i)
\end{align}
We retrain the softmax using by weighting the cross entropy cost of each image using (\ref{eq:sampleW}). After retraining, we run the BTS algorithm on the resulting beliefs and train an LSTM on the resulting action-values. We show the performance of the retrained LSTM as \emph{LSTM-i2} in figure \ref{fig:LSTMi2}. The retrained LSTM achieves a significant improvement over LSTM. We can repeat this procedure and see how much improvement can be achieved. In practice, we observed that the performance of LSTM starts to decline after the second iteration. The performance of retrained LSTM are shown in table \ref{table:perf}. 
}

\begin{figure}[ht!]
\centering
\includegraphics[width=0.35\textwidth]{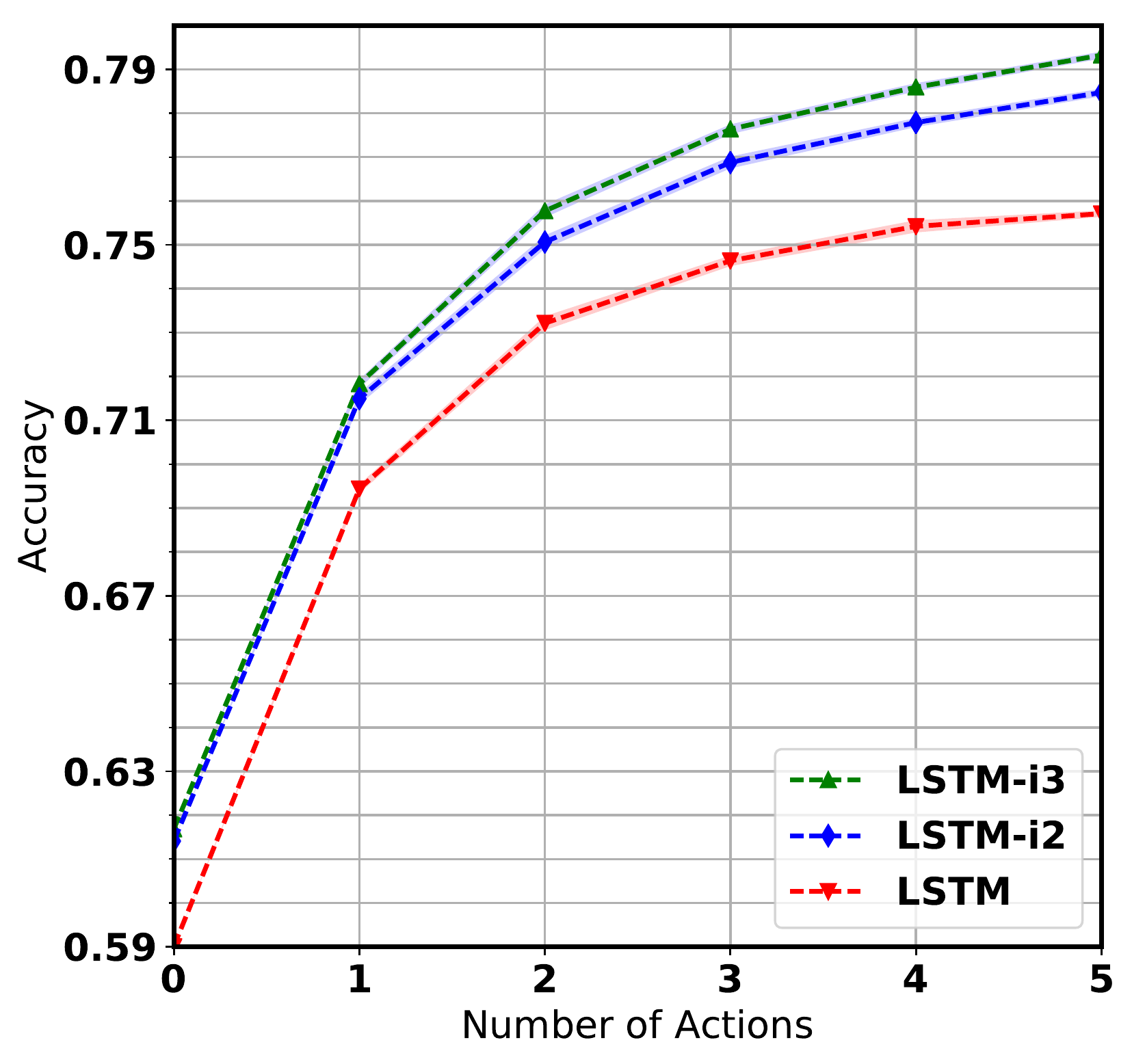}
\caption{comparison of the retrained observation function, and the corresponding supervised LSTM model with the original LSTM.}
\label{fig:LSTMi2}
\end{figure}

\section{CONCLUSIONS}
\par{
Active object recognition has received little attention from mainstream computer vision and machine learning communities despite its potential for improving recognition performance. Progress on AOR has been slow due to reliance of the AOR models on semi supervised or heuristic methods for object inspection policy discovery. In this work, we proposed a method that learns the object exploration policy in a supervised manner using training data. The proposed method has desirable properties, for example it is very fast at the test time and can generalize to novel objects since it does not require a generative model of objects. 
}
\par{
Recently there has been a large interest in attention models mostly for reasons besides object recognition. These models rely on reinforcement learning \cite{NIPS2014_5542} or variation inference \cite{ba2014multiple} to guide the system to discover suitable policies. In contrast, our proposed benefits from reduced training time and is free of the local optima that is a large problem in these approaches. By reducing the optimal policy inference to a supervised learning problem, one can use recent advances in supervised visual recognition to for learning policies for test data. 
}
\par{
We developed a weighting scheme for training of single images, that emphasizes images that result in higher value during exploration. The weight of each image denotes how useful is this image in achieving correct classification in the future.  Such weighting scheme potentially reduces overfitting of the single image classification model to images that have no discriminative information. Because of the complex background in GERMS images, there is high chance of overfitting to background cues during training. By performing BTS, the AOR has the opportunity to calculate the value of each image, and direct the single-image classification model to invest in more informative views of objects. 
}

%\addtolength{\textheight}{-12cm}   % This command serves to balance the column lengths
                                  % on the last page of the document manually. It shortens
                                  % the textheight of the last page by a suitable amount.
                                  % This command does not take effect until the next page
                                  % so it should come on the page before the last. Make
                                  % sure that you do not shorten the textheight too much.

%%%%%%%%%%%%%%%%%%%%%%%%%%%%%%%%%%%%%%%%%%%%%%%%%%%%%%%%%%%%%%%%%%%%%%%%%%%%%%%%

%%%%%%%%%%%%%%%%%%%%%%%%%%%%%%%%%%%%%%%%%%%%%%%%%%%%%%%%%%%%%%%%%%%%%%%%%%%%%%%%

%%%%%%%%%%%%%%%%%%%%%%%%%%%%%%%%%%%%%%%%%%%%%%%%%%%%%%%%%%%%%%%%%%%%%%%%%%%%%%%%
%\section*{APPENDIX}

%Appendixes should appear before the acknowledgment.

\section*{ACKNOWLEDGMENT}

The research presented here was funded by NSF IIS 0968573 SoCS, IIS INT2-Large 0808767, and NSF SBE-0542013 and in part by US NSF ACI-1541349 and OCI-1246396, the University of California Office of the President, and the California Institute for Telecommunications and Information Technology (Calit2).

\small
\bibliographystyle{ieee}
\bibliography{refs}

\end{document}